# Autoencoder Based Residual Deep Networks for Robust Regression Prediction and Spatiotemporal Estimation

**Lianfa Li[1,2,3], Ying Fang[1,2], Jun Wu [3], Jinfeng Wang[1,2]**

## ABSTRACT

To have a superior generalization, a deep learning neural network often involves a large size of training sample. With increase of hidden layers in order to increase learning ability, neural network has potential degradation in accuracy. Both could seriously limit applicability of deep learning in some domains particularly involving predictions of continuous variables with a small size of samples. Inspired by residual convolutional neural network in computer vision and recent findings of crucial shortcuts in the brains in neuroscience, we propose an autoencoder-based residual deep network for robust prediction. In a nested way, we leverage shortcut connections to implement residual mapping with a balanced structure for efficient propagation of error signals. The novel method is demonstrated by multiple datasets, imputation of high spatiotemporal resolution non-randomness missing values of aerosol optical depth, and spatiotemporal estimation of fine particulate matter <2.5 µm, achieving the cutting edge of accuracy and efficiency. Our approach is also a general-purpose regression learner to be applicable in diverse domains.

## 1. INTRODUCTION

Deep learning has achieved great successes in various domains including bioinformatics [1, 2], material science [3], reinforcement learning [4], computer vision, natural language processing and other domains [5] due to breakthroughs of a series of crucial techniques including backpropagation [6], fast graphics processing units [7], activation functions such as rectified linear unit (ReLU) and exponential linear unit (ELU) [8], convolutional neural network (CNN) [9], long short-term memory (LSTM) [10], generative adversarial network (GAN) [11], and deep belief network [12] etc. Many of these techniques are increasingly applied to remote sensing data analysis such as image preprocessing, pixel-level segmentation, target identification and extraction of semantic representation [13]. Recent examples include uses of CNN for target detections of ships and oil spills [14, 15] and classification [16, 17], LSTM for prediction of sea surface temperatures [18, 19] and target recognition [20], and GAN for hyperspectral image classification [21, 22] and segmentation [23].

One crucial aspect of deep learning is network depth [24]. Deeper networks have a larger number of parameters and much better generalization than shallow ones [25]. The earlier obstacle of vanishing or exploding gradient caused by deep hidden layers in artificial neural network (ANN) has been mostly addressed by efficient activations such as ReLU, normalization initialization and batch normalization with sufficient training samples available. Whereas the convergence issue can be solved by activation and normalization, added hidden layers may saturate and degrade accuracy quickly , as shown in many experiments of computer visions [26, 27]. Further, deep networks usually need a large training sample with substantial variations to find an optimal solution. Thus, training of deep networks with a small sample often results in non-convergence or degraded accuracy. For computer vision, residual connections [26] have been added continuously in sequence to a CNN, which largely reduces converging time and improves accuracy in prediction. But few efficient deep networks have been proposed for the regression of continuous variables with hardly any successful applications being reported.

In this paper, we propose a new architecture of autoencoder-based residual deep networks as a robust solution of deep learning for regression particularly of continuous variables. This method is broadly inspired by residual convolutional neural network in computer vision and recent findings of crucial shortcuts in the animal's brains in neuroscience [28, 29]. Residual vectors were are powerful shallow representation  for in image recognitions [30, 31]. Driven by this fact, residual CNN has been proposed to tackle the issue of accuracy degradation in deeper learning networks [26, 32]. In residual CNN, the shortcut of identity mapping is employed in a way of continuously stacked sequence (similar to ensembles of relatively  shallow networks [33]) to implement residual connections. Further, ANN is one crucial part of deep learning and vaguely inspired by the biological neural networks that constitute animal brains [34] . Recent findings show crucial roles played by hidden shortcuts in the brains for coordinated motor behavior and reward learning [28], as well as recovery from the damage [29]. Such shortcuts collaborate with regular neural connections to accomplish complex functionalities. Although the internal mechanism about the shortcuts in brains is unclear, such similar ideas of shortcut connections have been used in the domain of deep learning (e.g. residual CNN [32] and U-Net [35]).

[1] State Key Laboratory of Resources and Environmental Information Systems, Institute of Geographic Sciences and Natural Resources Research, Chinese Academy of Sciences, Datun Road, Beijing 100101, China
[2] University of Chinese Academy of Sciences, Beijing 100049,China
[3] Keck School of Medicine, University of Southern California, Los Angeles, CA 90032, USA
[4] Program in Public Health, College of Health Sciences, University of California, Irvine, CA 92697, USA
   Corresponding author: Lianfa Li (lspatial@gmail.com).

In the approach proposed, we introduce residual connections into the autoencoder-based architecture to implement the general-purpose residual deep learning. Different from residual connections stacked continuously in CNN [26], we take advantage of the balanced structure of encoding and decoding layers in the autoencoder, and leverage the shortcuts of identify mapping as residual connections from the shallow layers in encoding to their deep layers in decoding. Thus, forward and backward signals can be propagated directly in a nested way between an encoding layer and its decoding counterpart. Use of identity mapping is effective for propagation of error signals. We further use ReLU or ELU activation, which ensures speedy converging and effectiveness of error signals propagation.

To validate the proposed residual deep networks, we first tested the simulated dataset with a small sample size, and six benchmark datasets (3 for classification and 3 for regression) from the UCI repository of machine learning (http://archive.ics.uci.edu/ml).

We further applied the approach in real-world environmental science applications using remote sensing data. Remote sensing data have been widely used in classifying and categorizing geoscience information (e.g. land use and land cover, and target detection of geo-features) and in quantifying continuous environmental phenomena using regression models [e.g. the use of aerosol optical depth (AOD) for estimation of ground-level concentrations of fine particulate matter with diameter < 2.5 μm ($PM_{2.5}$)]. However, substantial non-random missing values of satellite-based AOD and the small number of ground-level $PM_{2.5}$ monitoring stations create challenges in applying deep learning such as CNN and multilayer perceptron (MLP) in remote sensing and environmental science. In this paper, we tested the proposed approach in 1) imputation of non-randomness missing values (mean proportion of missing values: 58%) for Multiangle Implementation of Atmospheric Correction Aerosol optical Depth (MAIAC AOD); and 2) surface $PM_{2.5}$ estimation in the Beijing-Tianjin-Tangshan area of China. To validate our approach, we conducted the independent tests using the AErosol RObotic NETwork (AERONET) to obtain the ground truth AOD, and the measured concentration from the $PM_{2.5}$ monitoring location of the US Embassy in Beijing.

We further examined the influence of network scales and structures (number of nodes for each hidden layer), activation functions, sizes of mini batches, and the inclusion/exclusion of spatial autocorrelation on regression model performance of simulated, AOD and $PM_{2.5}$ datasets. Finally, we proposed the optimal choice of the tested influential factors. Our proposed architecture is based on the antoencoder of the fully linked neural network as a general-purpose solution and can be applied in diverse domains.

## 2. RELATED WORK

**Autoencoder.** Autoencoder is a type of neural network with a symmetrical structure from encoding to decoding layers and the same number of input and output variables in an unsupervised or supervised manner [36, 37]. It aims to learn an efficient data representation (encoding) typically for dimensionality reduction. Fig. 1 presents a typical autoencoder with input, output, coding and hidden layers.

Assume a $d$-dimension input and output, **x**, weight matrix, **W,** bias vector, **b**, the set $\theta$ of parameters, the layer index, $L$, we have the following mapping formula:

$$\theta_{\mathbf{W,b}}(\mathbf{x}) : \mathbf{R}^d \to \mathbf{R}^d \qquad (1)$$

$$\theta_{\mathbf{W,b}}(\mathbf{x}) = f(\mathbf{W}^{(L)} f(\cdots f(\mathbf{W}^{(1)}\mathbf{x} + \mathbf{b}^{(1)}) \cdots) + \mathbf{b}^{(L)}) \qquad (2)$$

The parameters $\theta_{\mathbf{W,b}}$ can be obtained by the loss function between **x** and $\mathbf{x'}$ over the training data:

$$L = \frac{1}{2n}\left\|\mathbf{x} - \mathbf{x'}\right\|^2 = \frac{1}{2n}\left\|\mathbf{x} - \theta_{\mathbf{W,b}}(\mathbf{x})\right\|^2 \qquad (3)$$

Autoencoder provides a balanced network topology to implement the functionality similar to principal component analysis with mapping from encoding to decoding layers [38, 39].

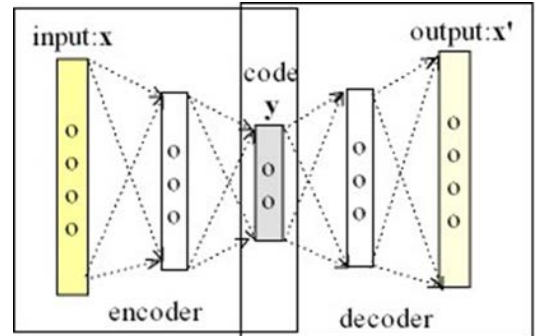

Fig. 1. A typical antoencoder with a symmetrical network topology and the same input and output.

**Residual learning in CNN.** Residual deep CNN has been proposed to tackle the issue of accuracy degradation in deeper learning networks [26, 32]. In this architecture, each residual unit includes several (e.g. two) continuous convolutional layers with batch normalizations and ReLU activations; all the residual units are stacked continuously to increase the depth and the generalization capability of the network (Fig. 3 of [26]).

Although residual CNNs have made breaking-through applications in computer vision, few studies on residual deep networks for regression are reported, likely due to the lack of optimal network topology and availability of dense sample data for deep network training.

**Spatial autocorrelation.** Spatial autocorrelation exists in most geoscience phenomena for surface estimations. Deep learning networks can not directly embed the structure of spatial autocorrelation. But spatial autocorrelation can be captured in deep network regression by using spatial coordinates and their derivatives (like square and product) as ones of the covariates. We may also employ the approach of nearest neighbors or kernel density to derive the layer of spatial nearest neighbors. Spatial autocorrelation, if embedded within the models, can improve the model's performance.

## 3. RESIDUAL DEEP NETWORK

**Autoencoder-based architecture.** Autoencoder (Fig. 1) is a mirrored network with encoding and decoding layers. For the decoding layers, every hidden layer may have a different number of nodes that introduce the variation for compression or adjustment of the dimensions. Autoencoder is a natural option for a residual regression network given that residual connections usually request the same number of nodes for the two (shallow and deep) layers considered to implement direct connection between them. Fig. 2 presents the architecture of a typical autoencoder-based residual deep network that has m inputs, k target outputs with symmetrical hidden layers plus the middle code layer between them. We also consider additions of activation functions and batch normalization to each hidden layer. We chose six symmetrical hidden layers for illustration purpose only. A larger network can be constructed similarly with the symmetrical network topology.

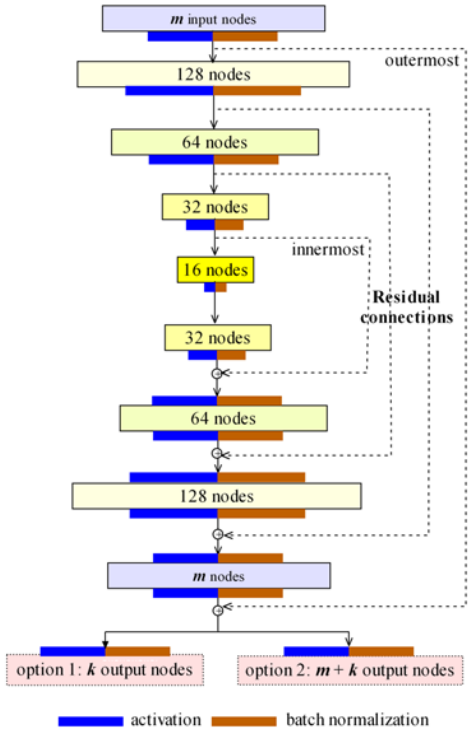

Fig. 2. Architecture of a typical autoencoder based residual deep network (m input, m output with symmetrical hidden layers with changes of the nodes number)

There are two options for k target variables, **y** to be output:

1) The target variables can be treated as the independent output layer that is fully linked with the autoencoder's output layer (option 1 in Fig. 3). This option makes the proposed network to have deeper layers with the inner topology to be the autoencoder. We can define the loss function ($\ell_O$) of mean square error (MSE) for regression or cross entropy for classification as the following:

$$L(\theta_{\mathbf{w,b}}) = \frac{1}{N} \ell_O\left(\mathbf{y}, f_{\theta_{\mathbf{w,b}}}(\mathbf{x})\right) + \Omega(\theta_{\mathbf{w,b}}) \quad (4)$$

where **y** is the observed values, $f_{\theta_{\mathbf{w,b}}}(\mathbf{x})$ is the predicted value, $\theta_{\mathbf{w,b}}$ represents the parameters to be optimized, and $\Omega(\theta_{\mathbf{w,b}})$ represents the regularization for $\theta_{\mathbf{w,b}}$ (L¹, L² or the others).

2) The target variables can be added to the autoencoder's output layer (option 2 in Fig. 3). This option introduces more interactions among the variables and more constraints on the target variables to be predicted. The loss function can be defined as:

$$L(\theta_{\mathbf{w,b}}) = \frac{1}{N}\Big[ \ell_O(\mathbf{y}, f_{\theta_{\mathbf{w,b}}}(\mathbf{x})) + \ell_{MSE}(\mathbf{x}, f_{\theta_{\mathbf{w,b}}}(\mathbf{x})) \Big] \\ + \Omega(\theta_{\mathbf{w,b}}) \quad (5)$$

The difference between option 1 and 2 lies in the placement of the target variables within the network. Comparison of Eq. (4) and (5) shows one constraint, $\ell_{MSE}(\mathbf{x}, f_{\theta_{\mathbf{w,b}}}(\mathbf{x}))$ on the parameters in terms of prediction of **y** in option 2. This constraint actually works as additional regularizers for **y** [25]. When sufficient samples are available, option 2 can effectively prevent over-fitting and boost the model's converging. But when only a limited number of samples are available, additional regularizers may make option 2 to have a high training error, thus option 1 is preferred then.

**Residual identity connection and implementation.** Shortcut connections have been added to neural network to address vanishing/exploding gradients [27,40] and degradation in accuracy in residual CNN [26,32]. We use the shortcut connection of identity mapping from the encoding layer to its corresponding mirrored decoding layer to implement residual networks in a nested way from the outermost to innermost layers (Fig. 2).

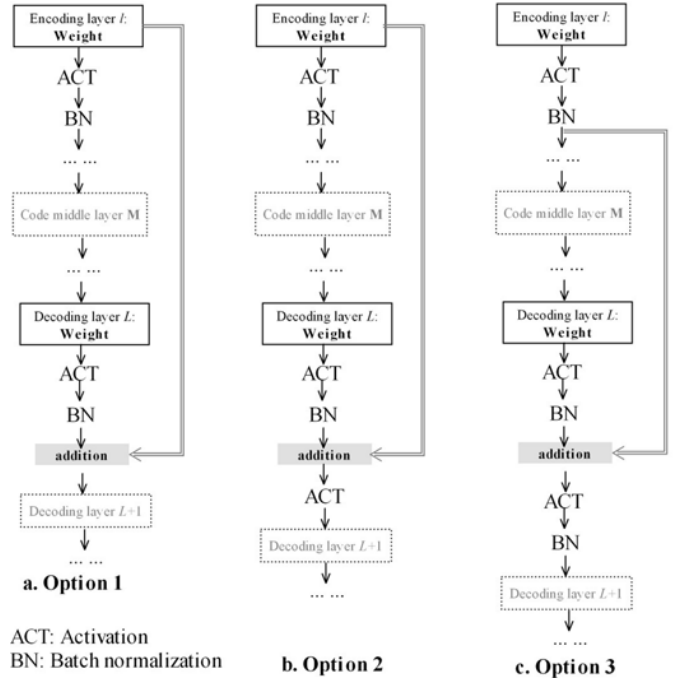

ACT: Activation
BN: Batch normalization

Fig. 3. Three options for a residual shortcut connection. Depending on whether neither activation nor batch normalization, just activation, or both activation and batch

normalization are added after the outputs of the hidden layers, there are three options for a residual connection (Fig. 3).

Assume the middle layer, $M$, the decoding layer, $l$, and its mirror decoding layer, $L$. Given $x_l$ and $y_l$ to be input and output of the encoding layer, $l$, respectively, and $x_L$ and $y_L$ be input and output of the decoding layer, $L$, respectively. With addition of residual identity connection, we have:

$$\mathbf{y}_L = \mathbf{x}_l + F(\mathbf{x}_L, \mathbf{W}_L) \tag{6}$$

$$\mathbf{x}_{L+1} = f(\mathbf{y}_L) \tag{7}$$

Since $L$ is a deeper layer for $l$, we can rewrite (3) as:

$$\mathbf{y}_L = \mathbf{x}_l + F(\cdots f(\mathbf{x}_l, \mathbf{W}_l) \cdots) \tag{8}$$

According to automatic differentiation [41], we can obtain the general derivative of the loss function, $L$ for $x_l$ that is used in turn to compute the gradients for the parameters, $W_{l-1}$:

$$
\begin{aligned}
\frac{\partial L}{\partial \mathbf{x}_l} &= \frac{\partial L}{\partial f_L(\mathbf{y}_L)} \frac{\partial f_L(\mathbf{y}_L)}{\partial \mathbf{y}_L} \frac{\partial \mathbf{y}_L}{\partial \mathbf{x}_l} \\
&= \frac{\partial L}{\partial f_L(\mathbf{y}_L)} \frac{\partial f_L(\mathbf{y}_L)}{\partial \mathbf{y}_L} (1 + \frac{\partial}{\partial \mathbf{x}_l} F(\dots f_l(\mathbf{x}_l, \mathbf{W}_l) \cdots))
\end{aligned} \tag{9}
$$

where $f_L(\mathbf{y}_L)$ represents the activation function for $\mathbf{y}_L$.

If we use the residual connect of option 1 in Fig. 3-a (no activation after addition of the shortcut identity connection), we can get the simpler version for (9):

$$\frac{\partial L}{\partial \mathbf{x}_l} = \frac{\partial L}{\partial \mathbf{y}_L} \frac{\partial \mathbf{y}_L}{\partial \mathbf{x}_l} = \frac{\partial L}{\partial \mathbf{y}_L} (1 + \frac{\partial}{\partial \mathbf{x}_l} F(\dots f(\mathbf{x}_l, \mathbf{W}_l) \cdots)) \tag{10}$$

There is one constant term, 1 in $\frac{\partial \mathbf{y}_L}{\partial \mathbf{x}_l}$ of (9) and (10) that makes the error information of $\frac{\partial L}{\partial \mathbf{y}_L}$ (or $\frac{\partial L}{\partial f_L(\mathbf{y}_L)} \frac{\partial f_L(\mathbf{y}_L)}{\partial \mathbf{y}_L}$ in (9)) directly propagated to the shallow layer, $x_l$ without concerning any weight layers. Further, since $\frac{\partial}{\partial \mathbf{x}_l} F(\dots f(\mathbf{x}_l, \mathbf{W}_l) \cdots)$ is not always equal to -1 to cancel out the gradient, $\frac{\partial L}{\partial \mathbf{x}_l}$ for mini-batch learning. This property can reduce vanishing of gradient during backpropagation and subsequent degradation in accuracy. Thereby, such shortcut connections can improve the network's generalization in collaboration with regular connections of deep layers.

For option 2 and 3 in Fig. 3, if the activation function of ReLU or exponential linear unit (ELU) or linear unit (LU), or/and batch normalization are added [$f_L(\mathbf{y}_L)$ in (6)] for better nonlinear modeling, they can also guarantee the nice property aforementioned according to (6).

Based on the autoencoder-based architecture, we adopt the nested shortcut connections of identity mapping from the outermost to innermost layers for residual deep networks (Fig. 2) that is different from the shortcut connections stacked continuously in residual CNN [26]. In the nested structure, besides regular backpropagation, the error information is also first directly transferred in the outmost layers of shortcut connections (from the last to first layers), thus effectively compensating update of the first layers' gradients. Then, such backpropagation occurs from the second nested layers until the innermost layers with the least effect. In the stacked way in CNN, the error information may be directly backpropagated along a longer path of residual connections than the nested way. If the components of activations and batch normalization are added between residual connections, the transfer may be affected somehow.

The autoencoder's structure provides a balanced network topology so that the shallow layers have their mirrored counterparts of deeper layers in the decoding component. Thus, the autoencoder-based structure is a natural option for residual regression deep networks. The nested residual connections make the error information to propagate efficiently throughout the whole network.

We further develop an iterative version of the proposed residual deep network (Algorithm 1) for option 3 in Fig. 3. In this algorithm, a stack is used to store the shallow layers in encoding and then pops the last item for addition of the encoding and decoding layers.

---

**Algorithm 1**: Residual Deep Regression Network (RDRN)

**Input**: *nfea*: number of features, *nnode*: list of the numbers of nodes for each layer; *k*: number of target variables; *acts*: list of activation functions for each layer; *dropout*: rate of dropout.

**Output**: Model of Residual Deep Network.

**Parameter**: *id*: index for layer depth; *stack*: a stack to store the previous layers.

Generate the input layer, *inlayer* according to *nfea*
Set *tlayer=inlayer*
**for** $i$, _ in enumerate(*nnode*) **do**
    Add a fully-linked layer with *nnode*[*i*] nodes to *tlayer*
    Add activation or/and batch normalization to *tlayer*
    **if** $i$<(length(*nnode*)-1) **do**
        Push *tlayer* to *stack*
    **else**
        Add a *dropout* layer to *tlayer*
    **end**
**end**
**for** $i$,_ in reversed(enumerate(*nnode*)) **do**
    Pop *player* from *stack*
    Add a fully-linked layer with *nnode*[*i*] nodes to *tlayer*
    Add activation or/and batch normalization to *tlayer*
    Add the addition of two layers: *tlayer* + *player* to *tlayer*
    Add activation or/and batch normalization to *tlayer*
**end**
Add the addition of two layers: *tlayer* + *inlayer* to *tlayer*
Add activation or/and batch normalization to *tlayer*
Add the output layer with *k* nodes (option 1) or *k+m* nodes (option 2) to *tlayer*
Return the model with input (*inlayer*) and output (*tlayer*)

---

In our proposed network (Algorithm 1 and Fig. 2), the basic building block consists of a shallow layer and its corresponding deeper counterpart with activations and batch normalization. This building block with optimal choice of its components is crucial for our approach.

**Table 1 Datasets for general test, imputation and spatiotemporal estimation**

| Dataset | | Domain | Target | Output type | #Features[a] | #Samples[b] |
|---|---|---|---|---|---|---|
| Simulated dataset | | Simulated | Regression | Continuous numerical values | 8 | 1000 |
| UCI datasets[c] | ADULT | Households | Classification | Binary | 123 | 5000 |
| | Heart disease (Cleveland) | Life | Classification | Binary | 14 | 303 |
| | Abalone | Life | Classification | Category (1-8, 9-10, 11)[d] | 8 | 4177 |
| | Combined cycle power plant | Energy | Regression | Continuous numerical values | 4 | 1030 |
| | Wine quality | Business | Regression | Continuous numerical values | 12 | 6497 |
| | Airfoil self-noise | Physical | Regression | Continuous numerical values | 6 | 1503 |
| MAIAC AOD | 365 sets of samples (each day for 2015) | Remote sensing | Imputation | Continuous numerical values | 17 | 34842-732294 |
| PM$_{2.5}$ | One dataset for 2015 | Environment | Spatiotemporal estimation | Continuous numerical values | 24 | 33118 |

*[a]*. Number of features (predictors). *[b]*: Number of samples. *[c]*: Datasets from the UCI benchmark repository of machine learning (http://archive.ics.uci.edu/ml). *[d]*. The target variable is classified to 3 categories according to the intervals.

For activations, we suggest the efficient activation functions of ReLU or ELU for most layers except the output layer. ReLU and ELU have identity function and the constant derivative of 1 for x>=0. Thus, they can partially keep efficient backpropagation of the error information. For x<0, ReLU is 0 but ELU has an exponential function ($\alpha(e^x - 1)$) with an exponential derivatives. With similar property for the positive input as ReLU, ELU can be a strong alternative to ReLU and it can well capture non-linear characteristics for negative input in practical applications [42]. For the output layer, we suggest the activation function of tanh or linear. The function of tanh that can better capture non-linearity than logistic activation with its symmetrical value range around the mean of 0 [43].

Batch normalization can also be added to each hidden layer to solve the issue of internal covariate shift [44] and to speed up the learning procedure in our architecture.

## 4. TEST

**Test datasets** In total, we examined nine datasets from four sources (Table 1), including one simulated dataset, six independent datasets from UCI benchmark repository of machine learning (three for classification and three for regression), MAIAC daily AOD of 2015 with 365 days for the Beijing-Tianjin-Tangshan area (Supplementary Section 1 and Supplementary Fig. 1), and 2015 ground PM$_{2.5}$ measurements with the covariates for the Beijing-Tianjin-Tangshan area (Section Methods gives more details). For each dataset, we randomly drew 20% of data for independent test, and then 20% from the rest 80% samples for validation (used to improving the models); the rest 64% samples were used to train the models. Given randomness of the learning algorithms, we used the averages of performance metrics over the training results of five times of each dataset for comparison.

**Table 2  Performance for general test, imputation and spatiotemporal estimation**

| Dataset (output type) | | Network structure | B. size[a] | Performance of training models | | | | Performance of model validation | | | | Performance of Independent test | | | |
|---|---|---|---|---|---|---|---|---|---|---|---|---|---|---|---|
| | | | | NORES[b] | | RES[c] | | NORES | | RES | | NORES | | RES | |
| | | | | RA[d] | RC[e] | RA | RC | RA | RC | RA | RC | RA | RU[f] | RA | RU |
| Sim. | One dataset (out1)[g] | [32, 16, 8, 4][h] | | 0.78 | 0.22 | 0.95 | 0.05 | 0.67 | 0.32 | 0.81 | 0.17 | 0.63 | 201 | 0.84 | 132 |
| UCI | ADULT (out1) | [256,128,64, 32,16,8] | 400 | 0.86 | 0.42 | 0.86 | 0.36 | 0.85 | 0.42 | 0.86 | 0.36 | 0.85 | 0.76 | 0.86 | 0.79 |
| | Heart disease (out1) | [128,64,32,1 6] | 16 | 0.78 | 0.46 | 0.91 | 0.26 | 0.70 | 0.56 | 0.83 | 0.50 | 0.79 | 0.78 | 0.83 | 0.80 |
| | Abalone (out1) | [128,64,32,1 6,8] | 96 | 0.68 | 0.71 | 0.72 | 0.62 | 0.64 | 0.72 | 0.66 | 0.73 | 0.65 | 0.55 | 0.66 | 0.55 |
| | Combined cycle power plant (out1) | [96,64,32,16] | 128 | 0.91 | 0.057 | 0.99 | 0.013 | 0.20 | 0.87 | 0.88 | 0.15 | 0.27 | 14.7 | 0.85 | 6.72 |
| | Wine quality (out1) | [96,64,32,16] | 1600 | 0.84 | 0.16 | 0.94 | 0.05 | 0.36 | 0.64 | 0.44 | 0.60 | 0.33 | 0.71 | 0.43 | 0.67 |
| | Airfoil self-noise (out1) | [128,96,64,4 8] | 128 | 0.89 | 0.10 | 0.93 | 0.07 | 0.90 | 0.10 | 0.92 | 0.09 | 0.90 | 2.08 | 0.93 | 1.86 |
| AOD | 365 datasets (each day for 2015) (out2)[i] | [75,50,25,12] | 200 | 0.79( 0.56-0.98)[j] | 1.7e-5[k] | 0.89( 0.75-0.99) | 5.0e-6 | 0.80( 0.58-0.98) | 1.5e-5 | 0.89( 0.76-0.99) | 4.7e-6 | 0.73( 0.34-0.73) | 0.013 | 0.87( 0.72-0.99) | 0.01 |
| PM$_{2.5}$ | One dataset for 2015 (out1) | [128,96,64,3 2,16] | 2000 | 0.88 | 0.08 | 0.97 | 0.03 | 0.74 | 0.46 | 0.89 | 0.11 | 0.72 | 36.9 | 0.88 | 24.0 |

*[a]*: Mini batch size. *[b]*: NORES, regular neural network with Autoencoder as its internal structure. *[c]*: RES, residual deep network proposed in this paper. *[d]*: RA: Performance metric: coefficient of determination (**R** squared) for regression and **A**ccuracy for classification; only the UCI sources has three datasets for classification, i.e. ADULT, Heart disease and Abalone are the datasets for classification. *[e]*. RC: normalized **R**MSE for regression and **c**ross entropy for classification. *[f]*. RU: regular **R**MSE for regression and **A**UC (defined as the area under receiver operating characteristic curve) for classification. *[g]*: out1 represent option 1 of output (defined in Fig 2). *[h]*: the digits in the square brackets gives in sequence the number of nodes for each layer in the decode component of the architecture proposed. *[i]*: out2 represent option 2 of output. *[j]*: mean(minimum value-maximum value). *[k]*: just the mean shown.

We further examine the influences of different placements (e.g. Fig.3 for three sample options) of the shortcuts of identity mapping, activation functions and batch normalizations on the generalization of the proposed residuals deep networks.

**Performance on all the datasets** The results (Table 2) show the residual deep network consistently and mostly considerably outperformed the one without residual connections [here regular (neural) network defined as that based on autoencoder but no residual connection for the purpose of comparison]: (1) the simulated dataset: an increase of 14% in $R^2$ and decrease of 0.2 in normalized RMSE in validation, and an increase of 21% in $R^2$ and decrease of approximately 34% in RMSE in independent test; (2) three classification dataset from the UCI's repository: an increase of 1-4% in accuracy in validation, and an increase of 0-3% in AUC (area under receiver operating characteristic curve) in independent test; (3) three regression datasets from the UCI's repository: an increase of 2-68% in $R^2$ and decrease of 0.02-0.72 in normalized RMSE in validation, and an increase of 3-58% in $R^2$ and decrease of 0.04-7.68 in RMSE in independent test; (4) MAIMC AOD: an averages increase of 9% in $R^2$ and average decrease of 1.03e-5 in RMSE in validation, and an increase of 14% in $R^2$ and decrease of 0.003 in RMSE in independent test (Fig. 4 showing distributions of $R^2$ and RMSE of the two networks); (5) $PM_{2.5}$: an increase of 15% in $R^2$ and decrease of 0.35 in normalized RMSE for validation, and an increase of 16% in $R^2$ and decrease of 12.9 $\mu g/m^3$ in RMSE in independent test.

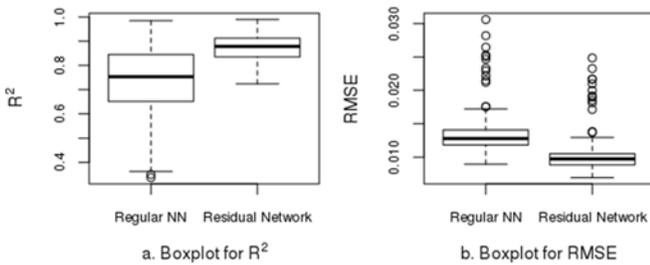

Fig. 4. Boxplots for $R^2$ (a) and RMSE (b) for daily network models of regular deep network vs. residual deep networks for MAIAC AOD imputations (N=365)

The training and validation loss and performance ($R^2$ for regression; accuracy for classification) along with increase of the training epoch are shown in Fig. 5. For MAIAC AOD, we just represented a typical day and the other days had similar trends. In total, residual deep network has better performance (lower loss and higher $R^2$) over regular networks along the training progress. Further, the results show that residual deep network more quickly converged to the optimal solution than regular network, illustrating its higher efficiency in finding a solution. For imputation of MAIAC AOD, the regular networks did not converge on 28 days.

The scatter plots of the simulated vs. predicted values are presented for the simulated dataset (Supplementary Fig. 2), AOD imputation (Supplementary Fig. 3) of the selected day and $PM_{2.5}$ estimation (Supplementary Fig. 4). The result shows that the residual deep network had less overestimation at low

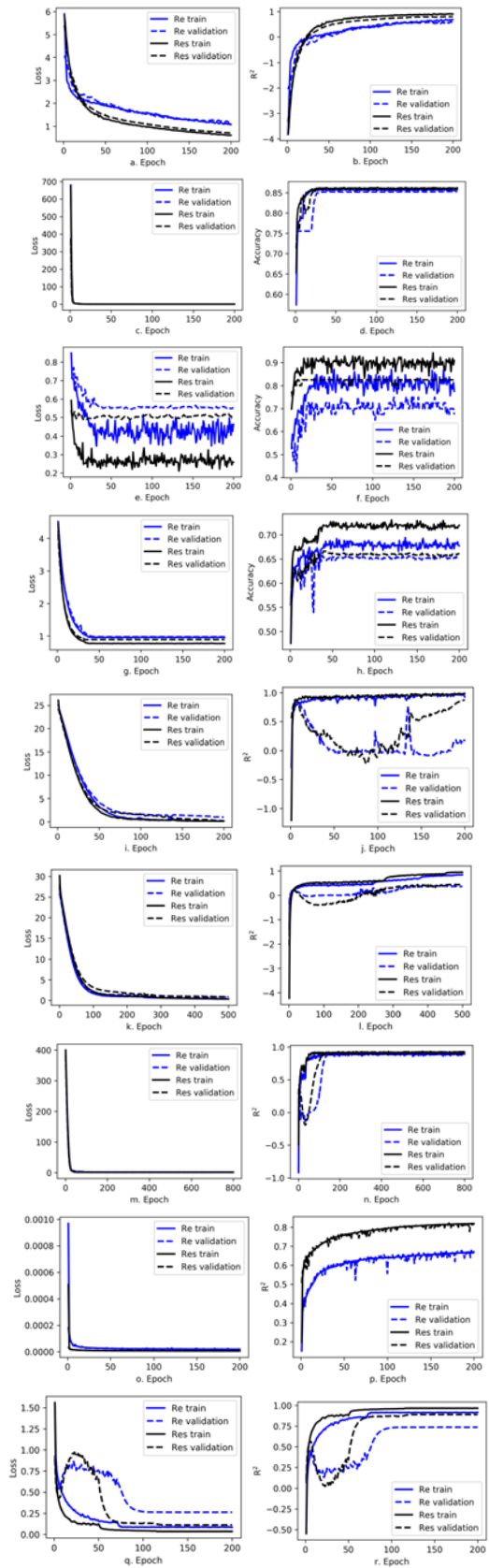

Fig. 5 Training curves of the loss function (a, c, e, g, i, k, m, o and q), and performance ($R^2$ for regression: b, j, l, n, p, r or accuracy for classification: d, f, h) for the simulated dataset (a-b), UCI dataset of machine learning (c-n), AOD imputation (o-p) and $PM_{2.5}$ estimation (q-r).

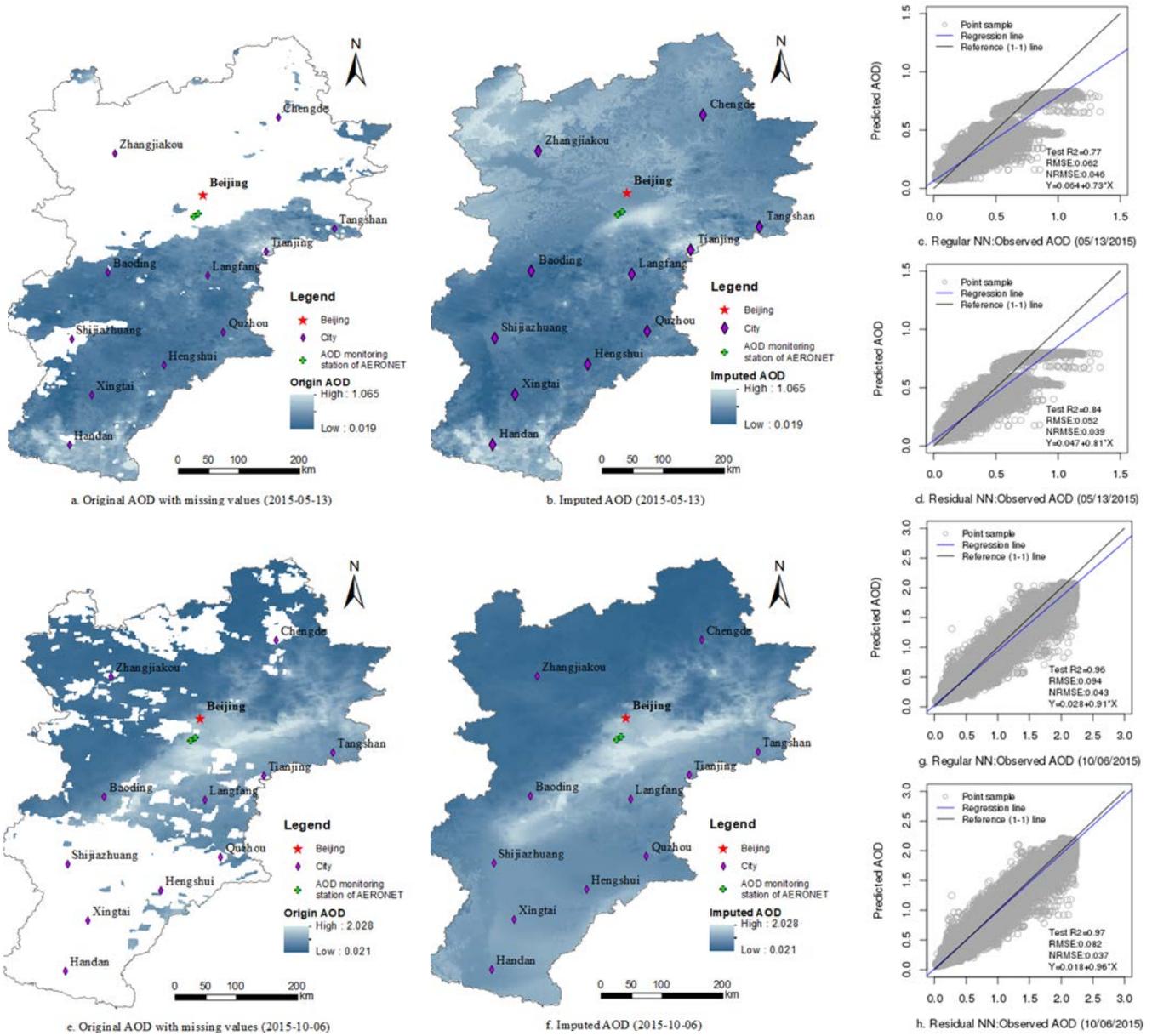

Fig. 6  The Beijing-Tianjin-Tangshan's MAIAC AOD surfaces of the original missing values (a and e) and the complete values (missing values imputed) (b and f) for two typical days of spring-summer (05/13/2015) and autumn-winter (10/06/2015) of 2015 with the plots of observed vs. predicted AOD for regular NN (also based on the symmetry structure of autoencoder, c and g) and the proposed residual deep network (d and h).

values and much less underestimation at high values than the regular network.

**Imputation of MAIAC AOD**   Residual deep networks improved $R^2$ by 13% over the regular networks with less underestimation at high AOD values for the regular network. The residual deep network had much better performance in prediction (less bias) than regular network.

We used the residual deep network to impute massive missing values of daily MAIAC AOD for 2015 in the Beijing-Tianjin-Tangshan area. Fig. 6 shows the maps of original AOD with massive missing values (a and e) vs. imputed AOD (b and f) on two typical days: a spring-summer day (05/13/2015) and an autumn-winter day (10/20/2015) in the Beijing-Tianjin-Tangshan area. Due to approximately 50% missing AOD, we selected the two typical days with different seasons. For the day of 05/13/2015, residual deep network improved $R^2$ of regular network by 7% with a decrease of RMSE by 0.01 (Fig. 6-c, d); for that of 10/20/2015, residual deep network improved $R^2$ of regular network by 1% with a decrease of RMSE by 0.012 (Fig. 6-g, h).

AOD is usually affected by multiple complex factors including artificial or natural emission sources, meteorology and elevation etc. Daily-level imputation of MAIAC AOD allows for daily variation of the associations between the predictors and target variable, and spatial autocorrelation which is beneficial for imputation of missing values [45]. Further, daily-level training dataset is convenient to implement for limited computing resources given the massive size of the images. Our results show that the residual deep network more reliably imputed missing AOD values with smooth variations in space. Our approach worked well to capture spatial variability of MAIAC AOD. The evaluation shows accurate match between our predicted missing values and the ground truth AOD from two AERONET stations (Supplementary Fig. 1 for their locations, and Supplementary Fig. 5 and 6 for the results): Pearson correlation of 0.93 for the 356 days of 2015 with statistical significance (p-value<2.2e-16).

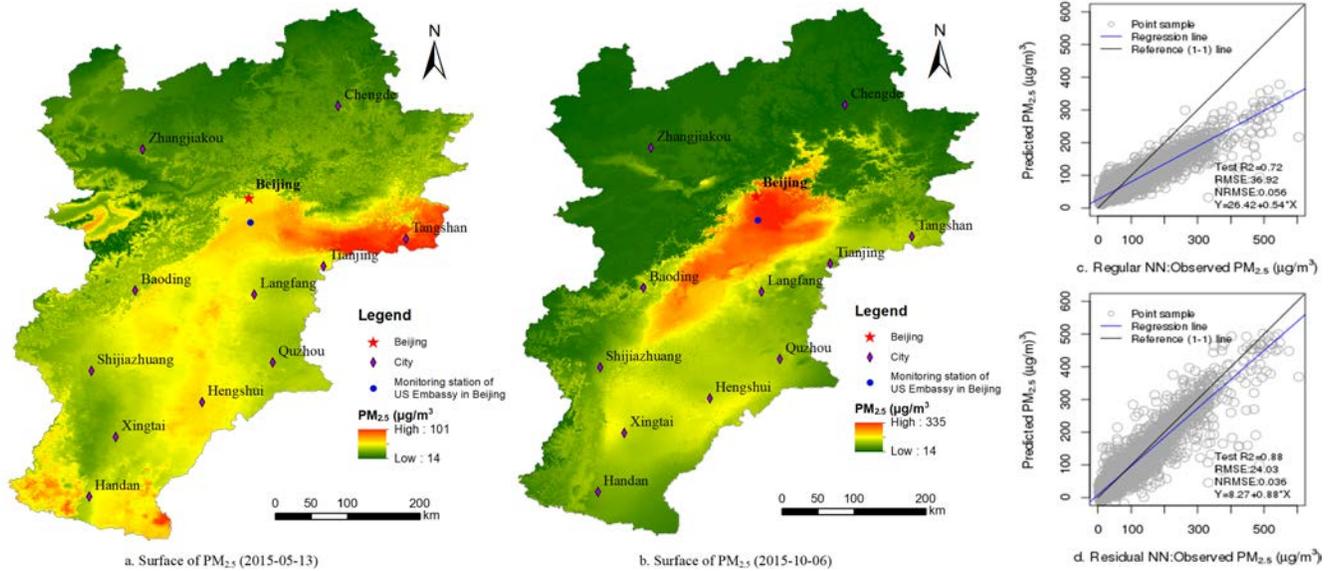

Fig. 7 The Beijing-Tianjin-Tangshan's PM$_{2.5}$ interpolation surfaces for two typical days of spring-summer (05/13/2015, a) and autumn-winter (10/06/2015, b) with the plots of observed vs. predicted AOD for regular NN (c) and the proposed residual deep network (d) (performances metrics such as R$^2$, RMSE, normalized RMSE, regression equations given also).

**Spatiotemporal prediction of PM$_{2.5}$**  The plots of the observed vs. predicted PM$_{2.5}$ (Fig. 7-c and d) show substantive underestimation at higher values for the regular network. The residual deep network had much better performance in prediction at higher values (less bias) than the regular network (R$^2$ improved by 16% and RMSE decreased by 12.89 μg/m$^3$).

The residual deep network was used to make the surface estimation of PM$_{2.5}$ at high spatial (1 km) and temporal resolution (daily) in Beijing-Tianjin-Tangshan area, one of the most heavily polluted areas in China. Given sparse monitoring data of air pollutants, high-accuracy surface estimation of PM$_{2.5}$ is important to identify local-scale emission, control air pollution, and examine health effects of air pollution. Our approach achieved the state-of-the-art accuracy (for independent test, R$^2$: 0.89; RMSE: 24 μg/m$^3$) based on a comparison to the comprehensive review of the related literature [45].

The daily predicted PM$_{2.5}$ surfaces (spatial resolution: 1km) on two typical days, the spring-summer day of 05/13/2015 and the autumn-winter day of 10/06/2015 in Beijing-Tianjin-Tangshan area (Fig. 7) show different patterns of spatial distribution of PM$_{2.5}$ between the two seasons: (1) higher PM$_{2.5}$ concentration in autumn-winter than that in spring-summer; (2) higher PM$_{2.5}$ concentration in eastern region in spring-summer but higher PM$_{2.5}$ concentration in middle region in autumn-winter.

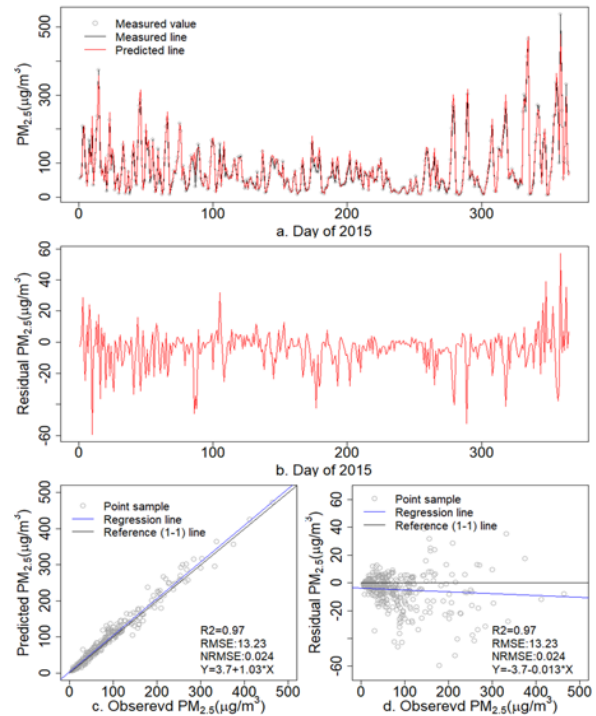

Fig. 8 The 2015-day time series of the observed and predicted PM$_{2.5}$ (a) and the residual PM$_{2.5}$ (b) with the scatters (c) of the observed vs. predicted PM$_{2.5}$ (c) and residual PM$_{2.5}$ (d) for the monitoring station of US embassy in Beijing.

With the monitoring sample from the US embassy in Beijing, independent validation showed excellent agreement between the monitoring samples and the predicted daily PM$_{2.5}$ (Fig. 8-a, c): Pearson's correlation:0.99 with p-value <2.2e-16. The time series (Fig. 8-b) and regular residuals (Fig. 8-d) of predicted PM$_{2.5}$ were checked for discernable spatial or temporal patterns. The time series residual plot showed higher residual in winter than the other seasons. Except for this, the residual patterns presented random mostly, illustrating spatial and temporal autocorrelation well captured by the models.

For spatiotemporal estimation of PM$_{2.5}$, our residual network incorporated spatial (coordinates and their derivatives) and temporal (day index) covariates to capture spatiotemporal variability of PM$_{2.5}$ concentration. Although spatial and temporal autocorrelations were not directly incorporated within the network, the evaluation showed high-accuracy estimates of PM$_{2.5}$. With a large number of parameters (49,195), the PM$_{2.5}$ residual network can well capture spatiotemporal variability of PM$_{2.5}$ with robust predictions (high R$^2$ and low RMSE).

# 5. PERFORMANCE ANALYSIS

Here are the results of performance analysis for network structure/complexity, activation functions, output nodes, mini batch size [25], and spatial autocorrelation [46] on the performance of the models.

### 1) Network Structure and Complexity

Network structure and complexity (number of hidden layers and number of nodes for each hidden layer) have significant influence on model performance. If the sample size is small, deep network may introduce over-fitting, lowering the accuracy in predictions. Residual connections can reduce degradation in accuracy as illustrated in the applications of CNN [26]. For our residual deep network, the results (Table 3) of the simulated, AOD and PM$_{2.5}$ datasets show similar positive effect of residual connections when the scale of the network increase. With the increasing scale and complexity (more parameters to be estimated) of the network, residual deep network had a smaller change (no or slight decrease) in the performance for independent test than the regular network. This indicates its insensitivity and robustness to the change in the scales of network.

**Table 3 Change in R$^2$ for Regular and Residual Autoencoders with Different Network Scale (Complexity)**

| Network scale[a] | Simulated Data | | MAIAC AOD | | PM$_{2.5}$ | |
|---|---|---|---|---|---|---|
| | nor[b] | res[c] | nor[b] | res[c] | nor[b] | res[c] |
| Small | 0.66 | 0.85 | 0.60 | 0.80 | 0.67 | 0.85 |
| Moderate | 0.81 | 0.88 | 0.76 | 0.90 | 0.81 | 0.88 |
| Moderately large | 0.86 | 0.87 | 0.88 | 0.93 | 0.86 | 0.87 |
| Large | 0.76 | 0.88 | 0.87 | 0.93 | 0.76 | 0.88 |

Network scale[a]: defined according to the number of nodes for each hidden layer; increase in sequence from small, moderate to large; nor[b]: regular network; res[b]: residual deep network.

With the increase in the number of residual connections starting from the outermost to the innermost layer, R$^2$ gradually increased and RMSE gradually decreased (Table 4). This illustrates that fully nested residual connections can better improve the network's generalization than the partially nested connections. Residual connections do not increase the number of parameters and the complexity in computing is not substantially increased.

**Table 4 Change in R$^2$ and RMSE in the independent test with different numbers of residual connections**

| #Residual connections | Simulated Data | | MAIAC AOD | | PM$_{2.5}$ | |
|---|---|---|---|---|---|---|
| | R$^2$ | RMSE | R$^2$ | RMSE | R$^2$ | RMSE (μg/m$^3$) |
| 1 | 0.79 | 147.5 | 0.76 | 0.06 | 0.83 | 28.6 |
| 2 | 0.82 | 141.5 | 0.79 | 0.06 | 0.86 | 26.8 |
| 3 | 0.84 | 126.9 | 0.81 | 0.06 | 0.88 | 23.1 |
| Fully nested connections | 0.88 | 117.1 | 0.84 | 0.05 | 0.89 | 22.6 |

From the outermost to innermost residual connection, different residual connections have different effects on the performance. Sensitivity analysis showed the most contribution from the outermost residual connection, with a decrement toward the innermost residual connection.

The activation and batch normalization added after the addition of residual connections within the network were tested. The results showed that both improved R$^2$ by 2-3% for the simulated data but no effect on the PM$_{2.5}$ data. Depending on different goals and samples, exploration is necessary to obtain the optimal network structure.

### 2) Activation functions

The activation functions for the hidden layers also affected training and test performances. Our tests show that ReLU and ELU worked well as the activation functions of the hidden layers. Sometimes ReLU achieved better performance for the models than ELU (e.g. PM$_{2.5}$ estimation) but it was more sensitive with significant difference in the accuracy (R$^2$ and RMSE) between the training and test than ELU.

With the output's activation, linear function often worked similar to or better than the other activations including ReLU, ELU and tanh with higher efficiency in computing.

### (2) Output type

There are two options for the output type as shown in Fig.3. Addition of $m$ input variables to the output variables works as strong regularizers on the $k$ target variables to be predicted. Such regularizers can effectively prevent model over-fitting but may weaken the generalization of the models in training using a relatively small size of samples. Our tests (Table 2) demonstrate that 1) option 1 is the best output type for the simulated data and PM$_{2.5}$ data with a small sample size (1000 and 35,000 respectively); and 2) option 2 is the best output type for AOD imputation with a large sample size (generally over 100,000).

### (3) Mini batch size

Mini batch size is another important factor on training of the models. Grid search of mini batch size was conducted for an optimal value. The test results for the independent samples of the simulated and PM$_{2.5}$ datasets (Fig. 9) show the optimal mini batch size for best independent test R$^2$: 100 for regular (R$^2$: 0.74) and residual network (R$^2$: 0.88) of the simulated datasets; 32 for regular network (R$^2$: 0.63) and 128 for residual network (R$^2$:

0.80) of the MAIAC AOD datasets (just one day's result reported here); 64 for regular network ($R^2$:0.75) and 96 for residual network ($R^2$: 0.88) of the PM$_{2.5}$ dataset.

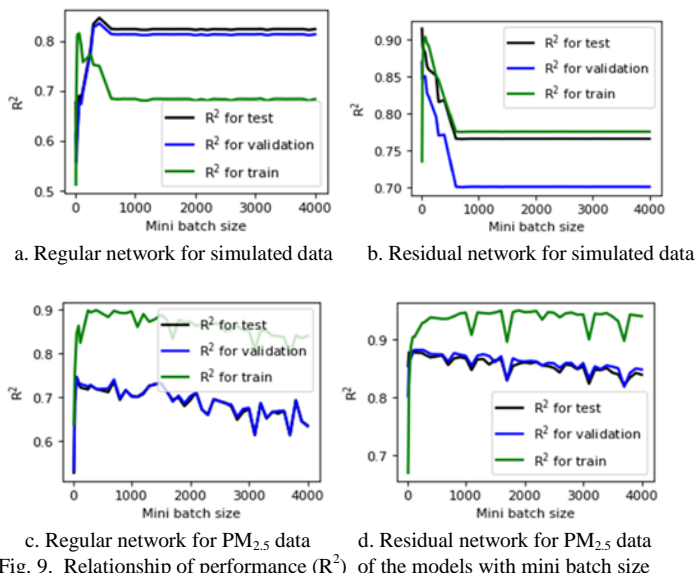

a. Regular network for simulated data    b. Residual network for simulated data

c. Regular network for PM$_{2.5}$ data    d. Residual network for PM$_{2.5}$ data

Fig. 9.  Relationship of performance ($R^2$) of the models with mini batch size

The result shows that different network topology may have different optimal mini batch size.

(4) Spatial Autocorrelation

Although the autoencoder-based architecture can't directly incorporate spatial autocorrelation within the models, the proxy variables such as coordinates can be incorporated within the dataset of regressors to capture the information of spatial autocorrelation. The results of PM$_{2.5}$ models with vs. without coordinate-related variables (Fig. 10) show about 6% lower $R^2$ and 5 μg/m$^3$ increased RMSE in the model without the covariates of coordinates in the independent test. The plots of the observed-predicted scatter points (Fig. 10-a vs. Fig. 7-d) and the residuals (Fig. 10-b) illustrated more underestimations over the high observed PM$_{2.5}$ for the models without spatial autocorrelation.

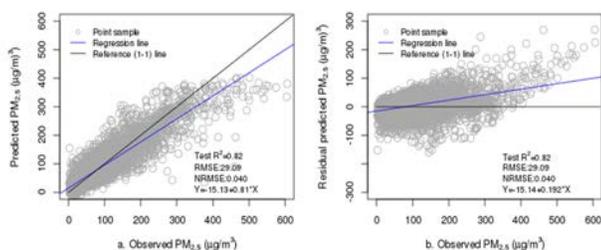

Fig. 10.  Scatter plot (a) of the observed vs. predicted PM$_{2.5}$ (b) and residual plot for observed PM$_{2.5}$ for the residual network without inclusion of the coordinates.

# 6. DISCUSSION

In this paper, we first proposed an autoencoder-based architecture of residual deep network. Autoencoder provides a balanced (encoding →decoding) structure based on which residual deep network can be naturally constructed. Such residual connections can compensate the loss in error backpropagation through the long path of regular deep layers, and collaborate with regular connections to reduce the known problem of degradation in accuracy.

By the tests upon the nine datasets with three hundred seventy-three models from four sources, i.e. one simulated dataset of a small sample size, six benchmark datasets form the UCI repository, three hundred sixty five daily imputations of non-random missing MAIAC AOD, and one estimation of PM$_{2.5}$, our approach consistently achieved the cutting-edge performance, with mostly (>80%) higher performance than regular network without residual connections, and less bias over extreme values at the two end of the continuous target variables (very low and very high values), which are not well predicted in typical regression models [47]. Further, the residual deep network is more stable for regression of continuous variables than regular network. The test in several daily-level MAIAC AODs showed inability of converging for regular network while residual deep network achieved reasonable performance for all the daily-level sets of AOD training samples.

Different from residual connections stacked continually in CNN, residual connections are embedded in a nested way within the proposed network for direct back-propagation of errors with a short path. The error information can be transferred the most efficiently along the outermost residual connection and then in a decremental way until the innermost residual connection. Our sensitivity analysis demonstrated this. The optimal network is that with all the nested residual connections available.

Performance analysis illustrated that the proposed residual nested network was robust to the scale and complexity of the networks. With increase of the network's scale, the network's performance is less affected by the training epochs. In our tests, residual deep network always converged with a reasonable prediction performance and speed, unlike regular network that did not converge for several AOD samples. For the others that regular deep network had a solution with lower performance, the sensitivity analysis showed that the performance was improved up to a similar level as residual deep network with more training epochs. This illustrates higher training efficiency (fast convergence) for residual deep network.

The proposed residual deep network is particularly useful for imputation of non-random missing data and high-resolution spatiotemporal surface estimations with a wide variety of applications in the relevant fields. With an unlimited number of parameters and reduction of degradation in accuracy with increased network complexity, our approach can be more efficiently employed to search the optimal solution. As demonstrated in practical applications of MAIAC AOD imputation and high spatiotemporal PM$_{2.5}$ predictions, our approach achieved optimal results with less bias, matching the ground truth values well. Although not directly embedding spatial and temporal autocorrelation within the framework, residual deep network could reliably capture spatial and temporal variability of the target variables with random patterns of the residuals.

It is important to incorporate spatial autocorrelation in the models. In the spatiotemporal estimation of MAIAC AOD and $PM_{2.5}$, the coordinates and their derivatives (square and product) were used as proxy variables to capture spatial autocorrelation. The other proxy variables to spatial autocorrelation include nearest neighbors of the target variable, the surface of kernel density and variogram in kriging. Nearest neighbors have been widely used in practical applications, including cellular automaton for land-use change [48, 49], and neural network for exposure estimation of $PM_{2.5}$ [50]. Nearest neighbors may produce sudden changes without smoothing for the target points with sparse distribution of the samples. Kernel density and variogram need specialist manual analysis to obtain reasonable parameters (e.g. bandwidth in kernel density, and sill, range and nugget in variogram). Use of the coordinates as proxy variables to spatial autocorrelation is relatively easy-to-implement and effective, as demonstrated in the practical applications. In the future, we can alter the network to incorporate the parameters from these methods for adaptive learning of their optimal solutions.

Besides, activation functions, batch normalization, network structure and mini batch size may affect model performance. Grid search can be conducted to optimize the configuration of these components for the best solution. The approach of autoencoder-based residual deep network is a general-purpose learner and can be also applied for diverse domains.

**Data availability**.   The six datasets from the UCI repository tested in this study are publicly available (https://archive.ics.uci.edu/ml). We shared the code, and the simulated and 80%-sampled $PM_{2.5}$ datasets in the python package, resautonet (https://github.com/lspatial/resautonet). The datasets of the MAIAC AOD and complete $PM_{2.5}$ in the Beijing-Tianjin-Tangshan area can be shared upon specific requests.

**Acknowledgments**   The study was supported by the Strategic Priority Research Program of Chinese Academy of Sciences (Grant No. XDA19040501), the Natural Science Foundation of China (Grant No. 41471376) and the opening project of Shanghai Key Laboratory of Atmospheric Particle Pollution and Prevention (LAP3).   We gratefully acknowledge the support of NVIDIA Corporation with the donation of the Titan Xp GPUs used for this research.

# Autoencoder Based Residual Deep Networks for Robust Regression Prediction and Spatiotemporal Estimation Supplementary Materials

Lianfa Li , Ying Fang, Jun Wu , Jinfeng Wang

## Supplementary Section 1.  Test Datasets and Validation

**Test datasets.**  We tested the proposed residual deep networks using nine datasets (Table 1) with three hundred seventy-three models trained from four sources: 1) one simulated datasets with a small sample size; 2) six datasets from the UCI benchmark repository of machine learning [51] (selected according to use frequency and data representative, and three for regression test, and three for classification); 3) daily MAIAC AOD derived from the MODIS satellites [52, 53]; and 4) daily $PM_{2.5}$ monitoring data in the Beijing-Tianjin-Tangshan area in China. The datasets aim to examine the convergence and performance of the proposed approach in training with a small sample size (simulated data and UCI benchmark datasets), imputations of large amount of non-randomness missing AOD, and spatiotemporal estimation of $PM_{2.5}$ using the samples from sparsely-distributed monitoring network of $PM_{2.5}$, respectively.

For the simulated dataset, we used eight random variables with uniform distributions of different scales and ranges plus a random noise with normal distribution (mean: 0, deviation: 100). In total, 1000 sample points were generated by the following non-linear formula:

$$y = x_1 + x_2 * x_3^2 + x_4 + \left(\frac{1}{x_5/500}\right)^{0.3} - x_6 + x_7^2 + x_8 + \varepsilon$$

where $\varepsilon$ is random noise, $\varepsilon \sim N(0,100)$.

For practical applications of the proposed approach in environment, we used the 2015 daily-level MAIAC AOD images and $PM_{2.5}$ measurement data that cover the Beijing-Tianjin-Tangshan area of China (Beijing as the capital) (Supplementary Fig. 1). As an urban agglomeration on North China Plain, this study region has over 41 million people with an area of about 196,068 $km^2$. It is a highly polluted area with $PM_{2.5}$ major sources including traffic emissions, industry etc.

We obtained the 2015 original images of daily 1- km resolution AOD with massive missing values from ftp://dataportal.nccs.nasa.gov/DataRelease (mean proportion of missing values for all the images: 58%). The regressors for imputation included meteorological parameters, MERRA2 AOD, elevation, coordinates and yearly average MAIAC AOD. The daily meteorological parameters included air temperature (°C), relative humidity (%), precipitation (kg/$m^2$), air pressure (hPa), and surface wind speed (m/s). We obtained the daily averages over the hourly meteorological data from China Meteorological Data Service Center (http://data.cma.cn/en) and also imputed for missing values using these monitoring data plus Reanalysis Data from Modern-Era Retrospective Analysis Version 2.0 (MERRA2, https://gmao.gsfc.nasa.gov/reanalysis/MERRA-2/). Three coordinate-related covariates ($x^2$, $y^2$, $xy$) were derived to capture spatial autocorrelation with $x$ and $y$. Since the study region is very large ( $km^2$) with the 1kmx1km resolution of MAIAC AOD, using all the 365 daily samples of 2015 to train a model is impractical due to high demands for memory and CPUs. Thus, we developed the daily-level models for imputation of missing values for time series MAIAC AOD: a model will be trained and tested separately for each day using the samples of three continual days with the middle day as the target day. The sample size (grid-days) ranged from over 34,842 732,294. The advantages of the daily-level models include allowing for daily variation of the association between the covariates and spatial autocorrelation, and implementability for limited computing resources.

For spatiotemporal surface estimation of 2015 daily $PM_{2.5}$ (unit: $\mu g/m^3$; spatial resolution: 1 km), we obtained daily $PM_{2.5}$ concentrations by averaging the observed hourly monitoring data from 92 monitoring stations (http://beijingair.sinaapp.com/) (Supplementary Fig. 1 for their locations). The regressors included daily AOD that include observed and imputed MAIAC AOD aforementioned, the meteorological parameters (the same data as those used in the imputation of MAIAC AOD), MERRA2 AOD, monthly NDVI from the MODIS MODND1D dataset (http://gscloud.cn), elevation, coordinates ($x$, $y$, $x^2$, $y^2$, $xy$). Compared with MAIAC AOD, the $PM_{2.5}$ dataset has a small sample size (about 33,181 sampling site-days) and a model can be trained for the whole dataset of 2015.

**Model validation**. For the simulated and UCI benchmark datasets, 20% of the samples were randomly sampled (without replacement) as the independent test data, and the rest 80% samples were further similarly sampled so that 80% of the rest samples were used to train the models, and 20% of the rest samples were used to validate and improve the models by feedback of each epoch training. For AOD imputation and $PM_{2.5}$ spatiotemporal estimation, we used the same sampling proportions for splits of the training, validation and test datasets, with the day index as the stratifying factor in sampling.

The results of the proposed residual networks are presented and compared with those of the autoencoder-based networks without residual connections. Network structure (number of nodes for each hidden layer) and output options are also listed in Table 2. Grid search [54] was conducted to obtain the optimal choices of network structure (numbers of nodes for the hidden layers and output types), activation functions, and the other parameters such as mini batch size and number of regressors etc. For optimal solution, option 1 (only $k$ target variables to be predicted in the output layer) was used for the simulated dataset and $PM_{2.5}$ interpolation due to their small size of samples; option 2 [$k$ target variables plus $m$ variables (same as the input variables) in the output layer] was used to impute MAIAC AOD given the big size of samples.

Further, since we created 365 networks in AOD imputation (one model per day for a full year of 2015), only the mean sample number, mean $R^2$ and mean RMSE are reported. The boxplots for $R^2$ and RMSE of daily networks of AOD imputation for regular network vs. residual deep networks are presented for a comparison in total performance of two networks.

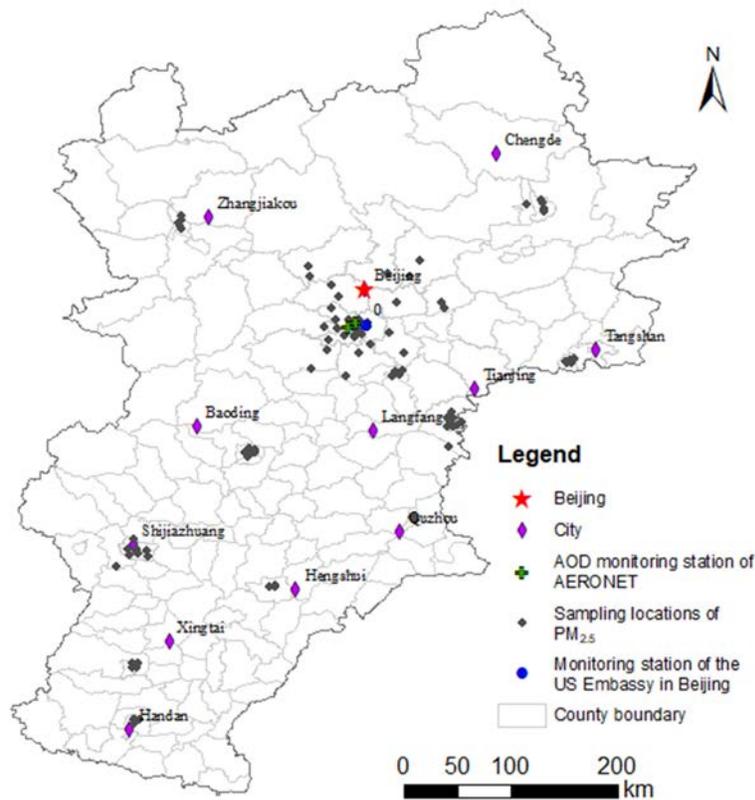

Supplementary Fig. 1.  Study region with the sampling locations of $PM_{2.5}$ and the $PM_{2.5}$ monitoring station of US embassy in Beijing.

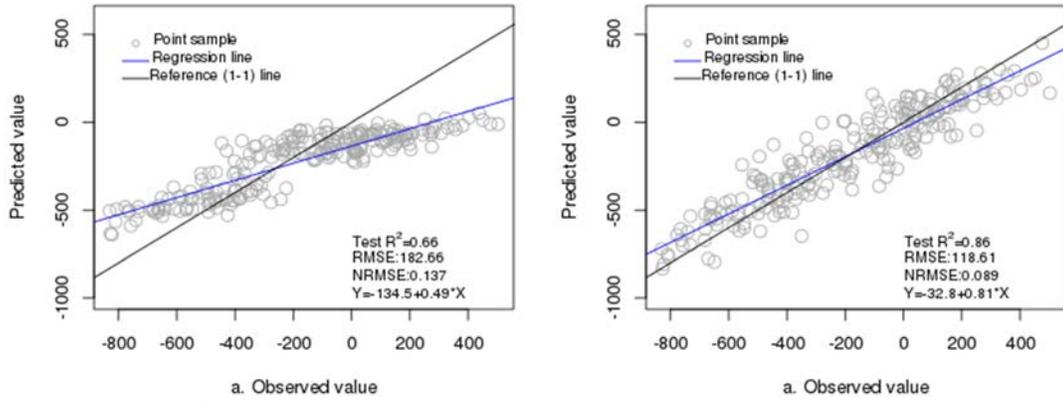

Supplementary Fig. 2. Scatter plot of simulated values vs. predicted values of the simulation for regular network (a) and residual deep network (b)

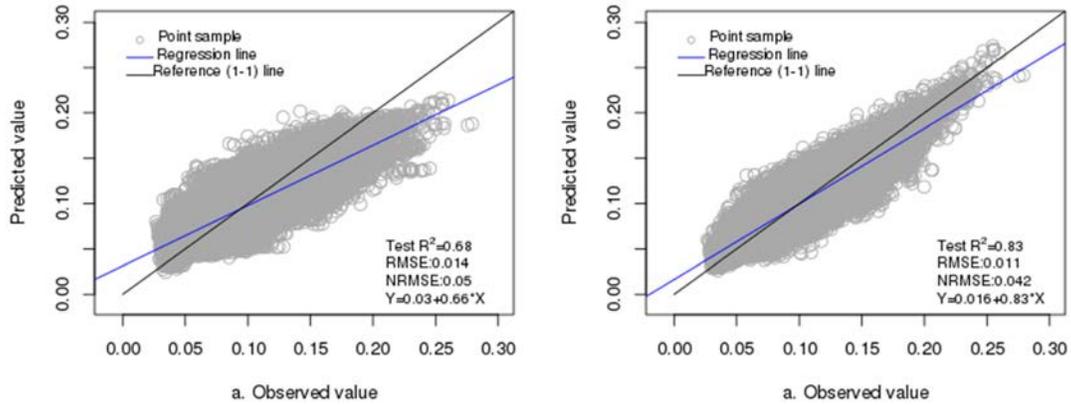

Supplementary Fig. 3. Scatter plot of observed vs. predicted MAIAC AOD for regular autoencoder (a) and residual regression network (b)

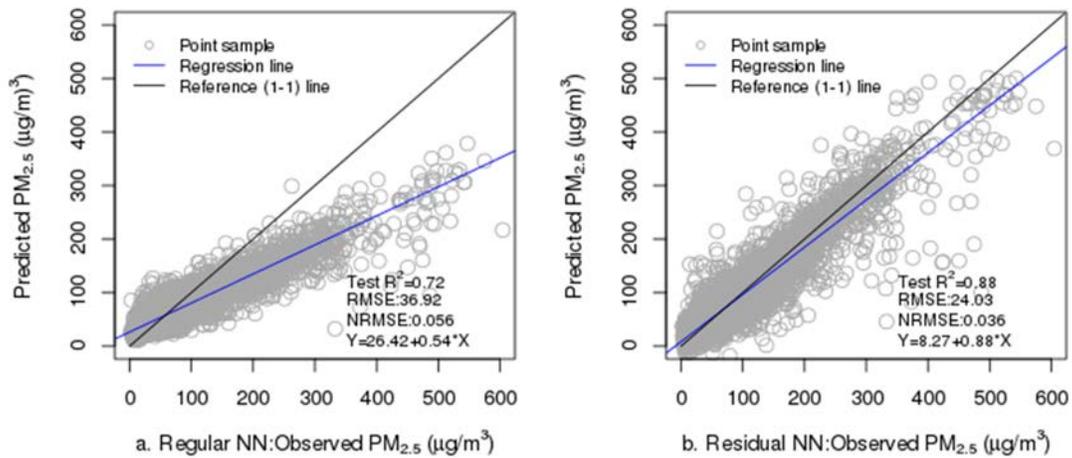

Supplementary Fig. 4. Scatter plot of observed and predicted $PM_{2.5}$ for regular network (a) and residual deep network (b).

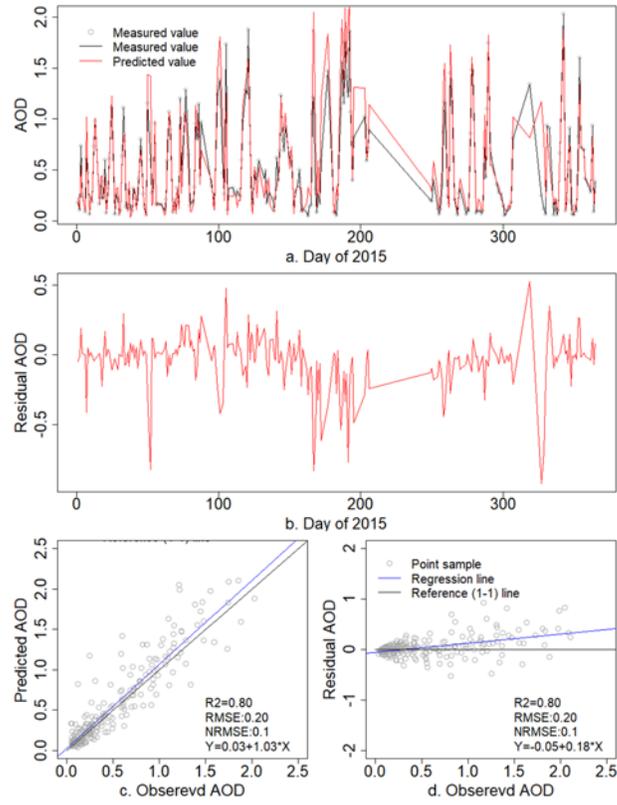

Supplementary. Fig. 5  Time series of observed vs. imputed daily AOD (a) and AOD residuals (b), scatter plot of predicted vs. observed AOD (c), and residual over observed AOD (d);  all based on independent test data from the AOD monitoring station of of AERONET in Chaoyang of Beijing.

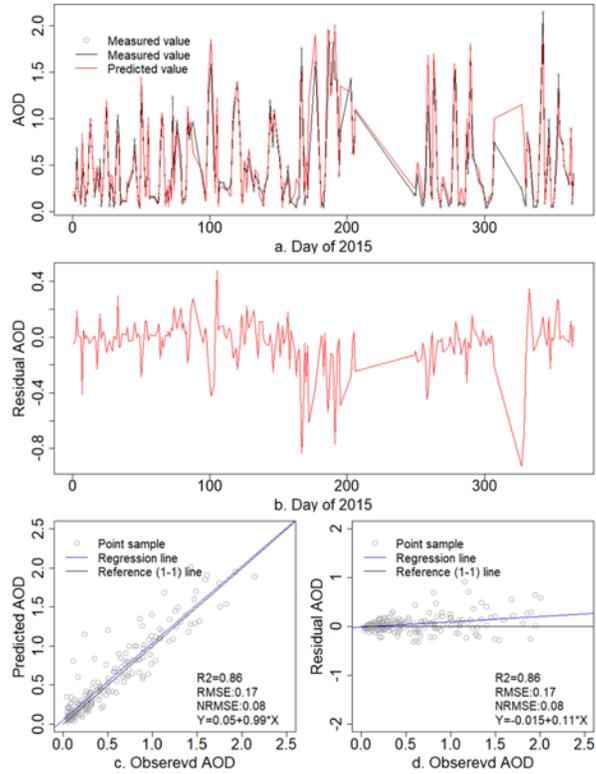

Supplementary. Fig. 6  Time series of observed vs. imputed daily AOD
(a) and AOD residuals (b), scatter plot of predicted vs. observed AOD (c),
and residual over observed AOD (d);  all based on independent test data
from the AOD monitoring station of of AERONET in Haidian of Beijing.